# Assessing Perceived Organizational Leadership Styles through Twitter Text Mining

La Bella, A., Fronzetti Colladon, A Battistoni, E., Castellan, S., & Francucci, M.

# Assessing Perceived Organizational Leadership Styles Through Twitter Text Mining

La Bella, A., Fronzetti Colladon, A Battistoni, E., Castellan, S., & Francucci, M

## Abstract

We propose a text classification tool based on support vector machines for the assessment of organizational leadership styles, as appearing to Twitter users. We collected Twitter data over 51 days, related to the first 30 Italian organizations in the 2015 ranking of Forbes Global 2000 – out of which we selected the five with the most relevant volumes of tweets. We analyzed the communication of the company leaders, together with the dialogue among the stakeholders of each company, to understand the association with perceived leadership styles and dimensions. To assess leadership profiles, we referred to the ten factors model developed by Barchiesi and La Bella in 2007. We maintain the distinctiveness of the approach we propose, as it allows a rapid assessment of the perceived leadership capabilities of an enterprise, as they emerge from its social media interactions. It can also be used to show how companies respond and manage their communication when specific events take place, and to assess their stakeholders reactions.

# Introduction

Leadership is a major issue in organizations and has gained more and more attention in the research literature (e.g., Bass, 2013; Behrendt, Matz, & Göritz, 2017; Tashiro, Lau, Mori, Fujii, & Kajikawa, 2011; Yukl, 1989). It has been shown that a good management of leadership skills can improve employees' engagement and satisfaction (Bass, 1985), as well as the positioning of the firm in the market, thus resulting in a higher level of organizational performance. Therefore, it is important for a firm to understand which leadership skills should be improved or better communicated to its stakeholders.

A favorable perception of a firm can be considered a strategic asset, as it can influence the access to resources and ultimately affect business performance (Van Riel & Fombrun, 2007). Accordingly, we maintain the importance for company managers to monitor the reputation of their firms also online, and we offer, in this study, a specific focus on perceived leadership styles.

Previous studies analyzed leadership-related constructs in online environments, for instance investigating the dimensions of leadership in online communities via automated text mining (Huffaker, 2010), suggesting leadership behaviors on Twitter for crisis management (Grubera, Smerekb, Thomas-Huntc, & Jamesd, 2015), or studying the association between leadership performance and social network positions (Tashiro et al., 2011). On a similar trend, we analyzed the perceived leadership styles of five large Italian multinational companies, by investigating the discourse about their brands and activities on Twitter. The Twitter dialogue was explored by means of a specific support vector machine (SVM), trained to automatically classify tweets according to the leadership dimensions defined in the work of Barchiesi and La Bella (2007).

We chose to analyze data from Twitter has it proved to be a valid media to engage stakeholders (Lovejoy, Waters, & Saxton, 2012) and because data are freely accessible via the Twitter API, with much less constraints than other social media platforms like Facebook.The text classification tool we propose in this paper can be useful for business leaders, to obtain a quick evaluation of the leadership styles of their company, for a specific timeframe, as emerging from the Twitter dialogue. Its strength mainly relies in the possibility to quickly collect and categorize huge amounts of text documents – tweets, in this case, but other sources are possible. Traditional interviews and surveys have limitations and can suffer from several biases (Podsakoff, MacKenzie, Lee, & Podsakoff, 2003), which can be partially overcome by the analysis of more spontaneous comments about a company on Twitter.

It is worth noting that, in our experiment, we were interested in assessing the perceived leadership styles of a company, as they might appear to the generic Twitter reader in a given timeframe. Accordingly, we did not perform a complete assessment of the communication and behavior of internal employees, or firm representatives, in comparison to external stakeholders. We analyzed the complete discourse about five companies, isolating the tweets which could be representative of specific leadership styles: for this reason, from now on we will always refer to leadership styles as they emerge from the Twitter discourse, as if a generic user would search a company name and read all the tweets in a timeframe. Thanks to our analysis, leaders could gain insights on whether their leadership style is suitable for the environment in which they operate or if there is a misalignment between real styles and what is discussed online. In Section 2, we provide a brief introduction on the main leadership theories and recall the ten factor leadership model on which our analysis is based. Section 3 describes the data collection process and the methodology used. Results from our case study are discussed in Section 4. In the last section, we discuss the managerial implications as well as the limitations of our method.

**Leadership and Leadership Styles**

Leadership is a wide concept which has attracted the attention of many scholars (Bass, 2013; Yukl, 1989). A review by Stogdill (1974) was already counting 3,000 leadership studies, with a great number of different approaches and perspectives in the exploration of the topic. More recently, Day and colleagues (2014) discussed methodological and analytical issues in leadership research, considering studies published over the past 25 years.

One of the best known taxonomies in this field is the one introduced by Burns (1978), who recognized two different types of leadership as opposites: transactional and transformational. The former deals with negotiation and exchange of something valuable between the leader and his/her followers to reach previously stated goals. This approach requires the leader to have the ability and the power to judge behaviors and consequently to bestow rewards or punishments (Waldman, Bass, & Einstein, 1987). Transformational leadership – on the other hand – is based on the leader's vision of the future that can inspire and motivate his/her followers: an effective vision should leverage on high ideals and values and never involve negative emotions, such as fear or the feeling of being threatened. Moreover, the leader interacts with others in a mutual beneficial way, so that each person – including himself/herself – can reach higher level of inspiration, motivation and self-empowerment (Bass, 1985; Conger & Kanungo, 1987; House, 1977; Podsakoff, MacKenzie, Moorman, & Fetter, 1990; Tichy & Devanna, 1986; Trice & Beyer, 1986; Yukl, 1989).

According to Burns' vision (1978), transactional and transformational approaches are mutually exclusive. Later studies (e.g., Bass, 1985) considered the possibility of an integration, with the leader adopting both styles at different times or under different circumstances. Bass (1985) recognized that both styles could be used to achieve goals, with one enforcing the other in such a way that a transformational leadership is likely to be ineffective in the total absence of the transactional style (Lowe, Kroeck, & Sivasubramaniam, 1996). In Bass' vision, a transformational leader produces an effect on his/her followers by making them aware of the tasks to be accomplished, of their importance, of the possible ways to achieve goals, and by activating their higher-order needs, making them transcend their self-interests to focus on those of the organization (Yukl, 1989). Transformational leadership is generally recognized as a shared process: indeed, the leader influences followers to engage them in the mission of the organization; simultaneously, the followers can influence one another and influence the leader's behavior, as he/she faces resistance or positive responsiveness.

The transformational leadership theory differs from leadership theories based on charisma (House, 1977; Conger & Kanungo, 1987; Conger & Kanungo, 1998; Conger, 1989; Shamir, House, & Arthur, 1993). Bass (1985) considers charisma just as a component of the transformational leadership. The charismatic leader emerges as somebody who is trusted and respected – similarly to the transformational leader – with the additional attribution of some exceptional qualities or characteristics, which cause the followers to show an unquestioned acceptance of whatever the leader posits as a correct behavior. Influence has mostly a one-way direction: from the leader to the followers. Charismatic leaders generally pose an appealing vision, show virtuous behaviors, make self-sacrifices, take personal risks to reach the stated goals, are self-confident, and are able to catch the attention of the followers with unconventional actions (Yukl, 1989). They are more likely to emerge in situations of crisis – actual or evoked – when people feel blocked and need a guide. To be more specific, we summarize the main characteristics of the presented leadership theories in Table 1.

**Table 1**

Transactional, transformational and charismatic leadership.

| Leadership style | Model Components | Description |
| --- | --- | --- |
| Transformational leadership (Bass, Avolio, Jung, & Berson, 2003) | Idealized influence (charisma) | Leaders are admired, respected, and trusted. Followers emulate their leaders and identify with them. The leaders put their followers' needs over their own needs. The leaders share risks with the followers and behave in a way which is consistent with underlying ethics, principles, and values. |
| | Inspirational motivation | Leaders behave in ways that motivate those around them, giving a meaning to all the activities that they carry out. They set up challenging personal objectives. Individual and team spirit is aroused. The group shows enthusiasm and optimism. The leader encourages the followers to envision attractive future states. |
| | Intellectual stimulation | Leaders stimulate their followers' effort to be innovative and creative, by questioning assumptions, reframing problems, and approaching old situations in new ways. There is no public criticism of the individual members' mistakes. New ideas and creative solutions to problems are encouraged. |
| | Individualized consideration | Leaders pay attention to each individual need of the followers, acting as coaches or mentors. Followers are empowered. New learning opportunities are created within a supportive organizational climate. Individual differences, in terms of needs and desires, are addressed and recognized. |
| Transactional leadership (Bass et al., 2003) | Contingent reward behavior | Leaders clarify expectations and offer recognition when goals are achieved. The clarification of goals and objectives usually improve individual and group performance. |
| | Passive management by exception | Leaders either wait for problems to arise before taking action, or take no action at all, showing a passive avoidant behavior. Such passive leaders do not specify agreements, clarify expectations, or set up goals to be reached by the followers. |

| | Active management by exception | Leaders specify the standards for compliance, as well as what constitutes ineffective performance, and may punish followers for being out of compliance with those standards. This style of leadership implies closely monitoring for deviances, mistakes, and errors, while implementing corrective action as quickly as possible. |
|---|---|---|
| Charismatic leadership (Conger & Kanungo, 1994) | Environmental assessment | Leaders are able to see opportunities and constraints in the environment, in members' skills and needs, and in challenges to the status quo. |
| | Vision formulation | Leaders share an inspirational vision and are effective communicators. |
| | Implementation (personal risk and unconventional behavior) | Leaders are seen as people who assume personal risks and engage in unconventional behaviors to reveal their extraordinary commitment and uniqueness. |

As regards the assessment of leadership behaviors, there were several attempts to develop questionnaires. In particular, Bass (1985) developed the Multifactor Leadership Questionnaire (MLQ): through this instrument, subordinates can rate the frequency with which their leader uses transactional or transformational leadership behaviors. The MLQ was revised and extended several times (e.g., Bass & Avolio, 1990; Bass, 1996). With regard to charismatic leadership, Conger and Kanungo developed the C-K Scale (Conger & Kanungo, 1994; Conger & Kanungo, 1998; Conger J. A., Kanungo, Menon, & Mathur, 1997). Similarly, Shamir, Zakay, Breinin, and Popper (1998) developed a questionnaire that measures four behaviors involved in charismatic leadership.

Once aware of the leadership profiles within a business company, it is also possible to investigate their impact on organizational performance. From this point of view, several studies agreed in stating that transformational leadership can foster employees' satisfaction, motivation and performance (Bass, 1985; Bass, Avolio, & Goodheim, 1987, Lowe, Kroeck, & Sivasubramaniam, 1996). However, some discrepancies were found, and many weaknesses were identified in all the three theories. For example, Yukl (1999) highlighted an overemphasis on dyadic processes in transformational leadership: the major goal of this theory was to explain the leader's influence on the individual follower, ignoring his influence at a group or organizational level. Similarly, some important transformational behaviors are missing in Bass' (1996) theory and in the MLQ, such as the ability to inspire and empower the followers, to facilitate agreements about objectives and strategies, and to build group identification and collective efficacy (Yukl, 1999).

Moreover, the choice of classifying leadership styles only considering the relationship between the leader and the followers was criticized, and a more integrated approach, that additionally includes external assessments, was proposed by Barchiesi & La Bella (2007). A similar approach requires the involvement of the stakeholders in the evaluation process, considering leaders' behaviors in different and complex situations: indeed, it is very important to offer a view on the external perceptions, which can greatly differ from the internal assessments.

**The Ten Factors Leadership Model**

Capitalizing on the presented theories, Barchiesi & La Bella (2007) developed a leadership model which includes ten factors, classified in four different areas of action (Figure 1). Each factor is an important component of the leadership profile, which can be assessed both at the individual and at the firm level. Each factor pertains to one or more of the following areas: the Symbolic area, related to the use of symbols, the inspiration of stakeholders and the sharing of a vision; the Behavioral area mainly focused on knowledge sharing skills and on the ability to manage human resources; the Political area, concerning the political power and the negotiation skills of the leader; the Structural area, related to the skills needed to maintain a functional structure of the organization and to effectively respond to external stimuli. According to the authors' point of view, each leader's behavior is reflected in a particular combination of the ten factors, which can vary over time. Although experience shows that a full mastery in each of the ten factors is difficult to achieve – and therefore rare – a good leader should be able to revise his/her use of the factors while acting change.

A questionnaire has been developed to evaluate the strength of a single leader or of a company for each of the ten factors (Barchiesi & La Bella, 2007). This questionnaire should be administered both to employees and external stakeholders, who can express judgments on a five-point Likert scale.

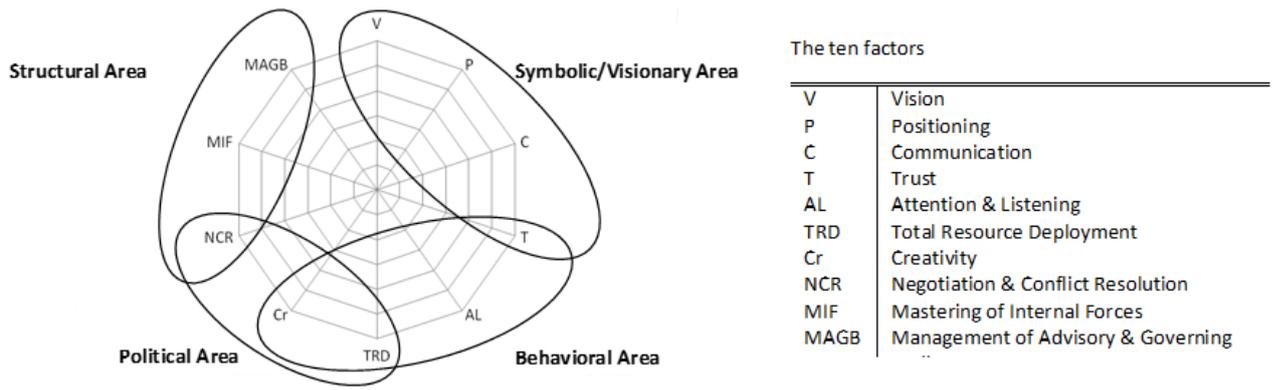

**Figure 1**. The ten factor leadership model.

Accordingly, a Leadership Index (LI) can be calculated to determine the intensity of leadership for a person or for an organization, related to the four areas of the model. The higher the value of LI, the higher the leadership skills.

$$LI = 10 \sum_{i=1}^{10} \frac{v_i}{4}$$

In the LI formula (Barchiesi & La Bella, 2007), $v_i$ represents the score on the i-th factor of the model expressed on the Likert scale. Leadership profiles can also be explored to see if they are more focused on some of the four areas of action or if, on the other hand, they are more balanced (see the example in Figure2).

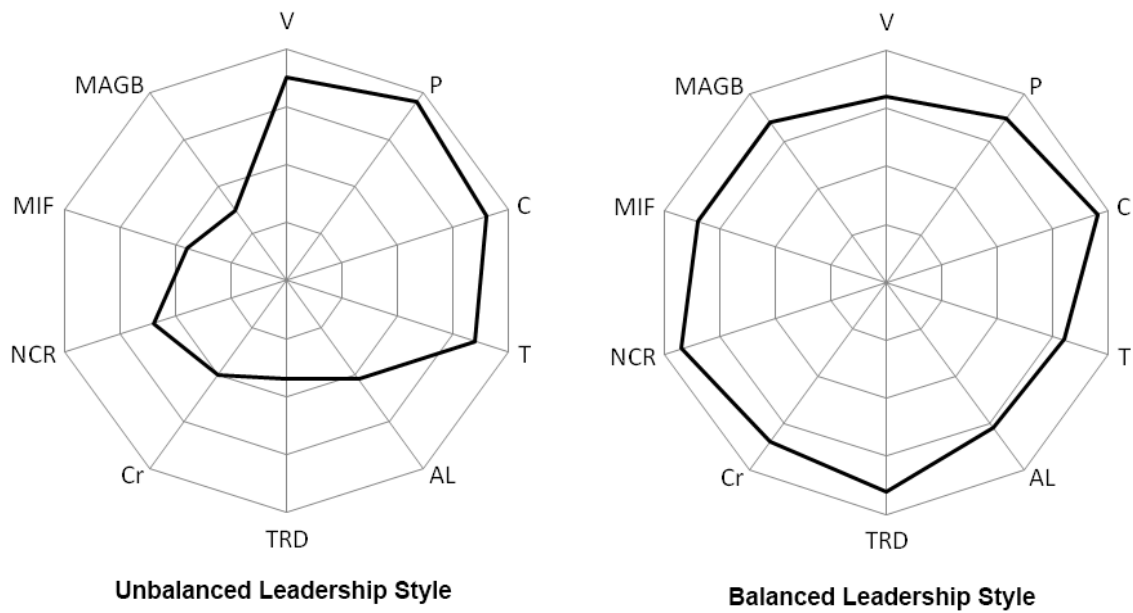

**Figure 2**. Balance in leadership styles.

Although the ten factor model proved to be effective in different business contexts and situations (Barchiesi & La Bella, 2007), it still relies on questionnaires. When questionnaires are collected by means of direct interviews, typically this process has several limitations, which span from the perceived costs for the interviewed – in terms of time spent and personal involvement for providing answers – to the actual costs and time spent to conduct the interviews, and to the risk of influencing the respondents' answers. On the other hand, if questionnaires are administered through some electronic procedure, the costs and the complexity of the process are reduced, but this could result in a lower response rate and in less accurate answers depending on the full understanding of the survey questions. In general, there are several limitations of the survey based approaches (e.g., Podsakoff et al., 2003), which can be partially overcome, using a text mining approach.

**A Big Data Approach to Assess Perceived Leadership Styles**

The use of a big data approach can provide a rapid access to a wide range of stakeholders and, at the same time, allows a more rapid assessment than the one provided by traditional survey tools. A quick assessment can indeed be vital for a timely and effective decision making process (De Mauro, Greco, & Grimaldi, 2016). In particular, the automatic detection of Twitter messages related to a specific topic relies on different well-known techniques (Farzindar & Wael, 2015). Among them, data mining and machine learning (Hastie,

Tibshirani, & Friedman, 2009; Murphy, 2012), natural language processing (Manning & Schütze, 1999; Jurafsky & Martin, 2009), information extraction and text mining (Hogenboom, Frasincar, Kaymak, & De Jong, 2011; Aggarwal & Zhai, 2012), and information retrieval (Baeza-Yates & Ribeiro-Neto, 2011). In our study, we used a text mining approach, applied to the four main areas of the ten-factors leadership model, following a research trend which already proved that online communication and behaviors can be linked to leadership profiles and performance (e.g., Grubera et al., 2015; Huffaker, 2010; Tashiro, 2011). On Twitter, we analyzed the discourse about five large Italian companies considering both the statements of their employees and the interactions with and among the stakeholders. Twitter already proved to be a valid media to engage stakeholders (Lovejoy, Waters, & Saxton, 2012; Rybalko & Seltzer, 2010). Besides accounting for everyday life stories, Twitter posts also report in real time local and global news and events. User-generated contents can provide access to valuable knowledge and actionable information (Farzindar & Wael, 2015). Consistently, companies are making use of Twitter for a variety of purposes: advertising and recommendation of products and services, forecasting of economic indicators, sentiment analysis of users' opinions about their products or those of competitors, increasing brand awareness and/or improving decision making and business intelligence (Elshendy, Fronzetti Colladon, Battistoni, & Gloor, 2017; Farzindar, 2012; Jansen, Zhang, Sobel, & Chowdury, 2009; Jiang, Yu, Zhou, Liu, & Zhao, 2011; Pak & Paroubek, 2010; Hollerit, Kröll, & Strohmaier, 2013). Twitter has also become one of the most used social media for official public relations, advertising, and marketing campaigns (Seltzer & Mitrook, 2007). In our context, statements of the stakeholders about a company can unveil perceived leadership styles and eventually draw the attention of managers on misalignments between perceived and expected styles. Our results can help companies to improve their online communication, thus fostering and creating the understanding and trust that are necessary to encourage others to follow a leader. "Leadership communication is the controlled, purposeful transfer of meaning by which leaders influence a single person, a group, an organization, or a community" (Barrett, 2006, p. 389).

## Methodology

Our case study started with the collection of tweets about the 30 Italian organizations in the 2015 list of Forbes Global 2000. Tweets were automatically collected over a period of three non-sequential weeks that span from December, 2015 to February, 2016. We used the Twitter fetcher included in the semantic and social network analysis software Condor[1], which allowed us to go back in time with the data collection up to ten days; in this way, for each week of data collection, we were able to fetch 17 days of tweets (for a total of 51 days analyzed). The total volume of collected tweets is 46,657, with an average number of 15,552.33 tweets per each block of seventeen days and a standard deviation of 5,165.54. All the collected tweets are in Italian. In a second phase, our attention focused on five out of the thirty companies, as they represented more than 90% of all the collected tweets in the first week of the analysis. These companies are Telecom Italia (now known as TIM), Eni, Enel, Unicredit Group and Pirelli & C.

In this experiment, we did not distinguish between tweets posted from CEOs, communication departments, or external stakeholders. Our aim was to provide a complete evaluation of the Twitter discourse about a company, regardless of the content sources, to assess the perceived leadership styles for a generic reader of Twitter. However, we removed from our sample a small proportion of tweets (less than 0.5%) coming from spammers (or Twitter bots). The detection of spammers – which can be automatized (Banerjee, Chua, & Kim, 2017; Zheng, Zeng, Chen, Yu, & Rong, 2015) – in this case has been carried out manually, according to the principles described in the work of Stringhini, Kruegel and Vigna (2010).

As our automatic classifier was not able to distinguish between positive and negative tweets in each leadership area, we carried out a preliminary sentiment analysis of the collected tweets, using the software Condor. From the analysis, it resulted that less than 1.5% of tweets had a negative sentiment. Being this number relatively small and almost evenly distributed (a little bit more for Eni, involved at the time in a journalistic inquiry), we decided to filter out these tweets. As a proposal for future research, we intend to refine our classifier, including an internal sentiment analysis of the tweets, with an algorithm trained on leadership-related semantic contexts.

As a second step of the analysis, we used the Natural Language Toolkit (NLTK) developed for the programming language Python 3 (Perkins, 2014), to pre-process the collected tweets, identify tokens, substitute capital letters, remove stop-words (such as conjunctions) and extract stems. Many stemming algorithms could be used; in this case study, we chose the Snowball Stemming, which is included in the NLTK package and is also available for the

---

[1] http://www.galaxyadvisors.com/products/

Italian language. The stemming of tweets helped us to reduce the number of features, i.e. to reduce the vocabulary, allowing a better accuracy of the classifier. To give a general example of the above-mentioned preprocessing steps, one could imagine to start with the following tweet:

*"At five thirty on Monday morning Luca was very relaxed"*

Once processed, the tweet would be transformed into the following list of words:

['five', 'thirti', 'monday', 'morn', 'luca', 'relax']

Following this procedure, we combined the pre-processed tweets to obtain a large matrix with as many rows as the tweets in our sample and as many columns as the words in the complete vocabulary extracted from the tweets. Each element $a_{ij}$ in the matrix had the value of 1 if the word represented by the column *j* was included in the tweet at the row *i*, and the value of 0 otherwise. Using this matrix in a machine learning algorithm would present the problem that each term would have the same discriminatory power in determining the class of a tweet. Indeed, there might be words that recur in almost every tweet, being almost useless for the classifier. To attenuate this effect, we transformed the matrix using the Inverse Document Frequency function (IDF) (Ounis, 2009).

Subsequently, we used a text classification technique to assign the collected tweets to five different clusters: one cluster for each of the four leadership dimensions – symbolic, behavioral, political, and structural – and a fifth one called "none of the others". A single tweet could also be assigned to multiple clusters. Indeed, the five are not mutually exclusive: tweets – even if their length is limited – can contain ideas, opinions and statements that can pertain to more classes. This is even more likely to happen as the four areas are in some parts overlapping. The only class which cannot coexist with the others is the fifth one. To start the supervised classification, we built a training set made of 3,000 tweets, randomly extracted from the collected ones – 1,000 tweets per collection period. The tweets in the training set were manually classified by three independent annotators and their agreement for each label has been measured by means of Cohen's Kappa. Values of Kappa ranged from a minimum of 0.77 to a maximum of 0.83, thus denoting a good level of inter-annotator agreement. The tweets that presented divergent annotations were reexamined by the three experts together, in order to converge on a single classification. To give an example, the tweets sharing the vision of a company or discussing it in a positive way were classified as tweets of the symbolic area;

on the other hand, tweets dealing with the creativity of a company have been classified with two labels (political and behavioral). Many algorithms are available for text classification – such as Naive Bayes (Lewis, 1998), Nearest Neighbor (Yang, 1999), Neural Networks (Ozmutlu, Cavdur, & Ozmutlu, 2008), Rule Induction (Apte, Damerau, & Weiss, 1994), and Support Vector Machines (SVM) (Vapnik, 2013). SVM were included among the most effective classification techniques, with the additional advantage of being more generalizable than others, such as decision trees (Wang, Sun, Zhang, & Li, 2006; Lee & Lee, 2006). The best choice in our case study has been to use a SVM algorithm with a Radial Basis Function (RBF) kernel. In order to train and execute the SVM classifier we used the package Scikit-learn which comprises several tools for data mining in Python 3 (Pedregosa et al., 2011). To solve the multi-label classification problem, we implemented a series of independent binary SVM classifiers, each one associated to a specific label. Therefore, each classifier defined whether a label was associated to a particular class or not. The fifth class has been filled in with the tweets that were not classified in one of the other four clusters. In this way, none of the tweets was left without a label. In addition, we tested a binary classifier for the "none of the others" cluster, to compare these results with the previous strategy. Results showed no significant differences. Each binary classifier was optimized separately – calibrating the parameters C and γ of the RBF kernel, as allowed by the Scikit-learn package (Pedregosa, 2011); subsequently, the classifiers were used to train the SVM for the multi-label classification. The final multi-label classifier showed good fit indexes, with a subset accuracy of 0.779, a micro-$F_1$ of 0.804, and a macro-$F_1$ of 0.741. As regards the binary classifiers of the five clusters, accuracy ranged from a minimum of 0.841 (cluster "none of the others") to a maximum of 0.942 (cluster "political") and $F_1$ ranged from 0.678 (cluster "behavioral") to 0.872 (cluster "none of the others").

**Companies included in the case study**

As previously mentioned, we selected the following large Italian companies, as they had the largest volumes of tweets in our data collection: Telecom Italia (now known as TIM), Eni, Enel, Unicredit Group and Pirelli & C. Here we provide a better introduction to them, together with some information which could partially relate to the study results.

      **Telecom Italia – TIM.**

Telecom Italia is a large telecommunications company operating in Italy and abroad. It was founded in 1994 by the merger of several state-owned telecommunications companies. In 1995 TIM was founded as a mobile telephony company which was later incorporated by Telecom Italia. A rebranding process, started in 2015, led the brand "TIM" to substitute the name of "Telecom Italia". Since 2010 Telecom Italia-TIM begun to provide customer service through social caring – characterized by a quick and direct interaction with customers over Facebook and Twitter. This effort led TIM to several acknowledgements: as an example, according to SocialBakers[2] TIM has been in the fifth position worldwide in terms of help replies for the fourth quarter of 2015 and for the first two quarters of 2016. Among the five analyzed enterprises, Telecom Italia-TIM and Unicredit Group are the only ones which received for 2016 the CRF Institute "Top Employers Italia" certification[3].

Telecom Italia-TIM constantly invests in new technologies, such as the development of a new 4G mobile network and of new high-speed fixed line connections. The investment strategy has been the object of many communication campaigns. In addition, these campaigns have been supported by corporate communication efforts with a high level of symbolic meanings, focused on organizational values. The corporate communication has also supported the rebranding process, which ended at the beginning of 2016.

### Eni.

Eni is an Italian multinational oil and gas company. Founded in 1953, it operates in 79 countries and is currently one of the top 20 world's largest industrial companies. Eni is a state-owned enterprise: the Italian government owns more than 30% of the company.

The number of tweets about the organization, collected during the first week of observation, is roughly four times the value obtained for the same firm in the other two collection periods. This is probably due to a journalistic inquiry that involved the company and to which the company decided to answer publicly via Twitter. In the inquiry, Eni was suspected to have paid a huge kickback to probe the seabed in Nigeria. The company made a big effort to control the discourse on Twitter, succeeding in significantly reducing the negative impact on its reputation.

### Enel.

Enel is an Italian multinational enterprise operating in the utilities industry. It produces and distributes electricity and gas in Italy and abroad. Established as a public company in 1962, it was privatized in 1999. At the present, the Italian government owns more than 25% of the company shares. Fortune acknowledged Enel with the fifth position in the "Change the world" 2015 list[4]. This list comprises companies that had positive social impact through activities that are part of their core business strategy. Enel achieved this acknowledgement thanks to its ability to overcome barriers to the development of renewable energy.

---

[2] https://www.socialbakers.com/free-social-tools/socially-devoted/q3-2016/
[3] http://www.top-employers.com/Certified-Top-Employers/?Certificate=61

### Unicredit Group.

UniCredit Group is an Italian global banking and financial services company founded in 1998, operating in 50 markets and in 17 countries. The Group is committed in social issues and is a reference point for nonprofit organizations, thanks to the Unicredit Foundation. The involvement in social topics is well promoted through social networks. Unicredit Group was rewarded with the 2016 CRF Institute "Top Employers Italia" certification[5], together with Telecom Italia-TIM. Layoff plans involving 12,000 employees were announced by Unicredit in September 2015.

### Pirelli & C.

Pirelli & C. is an Italian multinational tyre manufacturer, active in over 160 Countries with 19 manufacturing sites around the world. It is the world's fifth-largest tyre manufacturer, after Bridgestone, Michelin, Continental and Goodyear. Pirelli & C. experienced a change in the ownership of the company in 2015, when it was acquired by ChemChina (March-November 2015). Since 1907 the company has been sponsoring major sport competitions, becoming the exclusive supplier for the 2011-2019 Formula One Championships and for the FIM World Superbike Championship.

## Results

Table 2 shows the results of our classification algorithm, together with the number of collected tweets for each company.

**Table 2**

Volume of tweets and their classification for the five Italian companies.

| Organization | Week | Symbolic | Behavioral | Political | Structural | None of the others | Total |
|---|---|---|---|---|---|---|---|

---

[4] http://fortune.com/change-the-world/2015/
[5] http://www.top-employers.com/Certified-Top-Employers/?Certificate=61

| | | | | | | | |
|---|---|---|---|---|---|---|---|
| **Telecom Italia-TIM** | 1st | N | 498 | 312 | 126 | 266 | 844 | 1,813 |
| | | P | 27.47% | 17.21% | 6.95% | 14.67% | 46.55% | 100.00% |
| | 2nd | N | 63 | 183 | 16 | 47 | 656 | 938 |
| | | P | 6.72% | 19.51% | 1.71% | 5.01% | 69.94% | 100.00% |
| | 3rd | N | 104 | 103 | 78 | 54 | 735 | 974 |
| | | P | 10.68% | 10.57% | 8.01% | 5.54% | 75.46% | 100.00% |
| | Total | N | 665 | 598 | 220 | 367 | 2,235 | 3,725 |
| | | P | 17.85% | 16.05% | 5.91% | 9.85% | 60.00% | 100.00% |
| **Eni** | 1st | N | 1,709 | 799 | 843 | 512 | 5,493 | 8,102 |
| | | P | 21.09% | 9.86% | 10.40% | 6.32% | 67.80% | 100.00% |
| | 2nd | N | 187 | 144 | 73 | 316 | 1,107 | 1,827 |
| | | P | 10.24% | 7.88% | 4.00% | 17.30% | 60.59% | 100.00% |
| | 3rd | N | 176 | 71 | 256 | 202 | 1,576 | 2,023 |
| | | P | 8.70% | 3.51% | 12.65% | 9.99% | 77.90% | 100.00% |
| | Total | N | 2,072 | 1,014 | 1,172 | 1,030 | 8,176 | 11,952 |
| | | P | 17.34% | 8.48% | 9.81% | 8.62% | 68.41% | 100.00% |
| **Enel** | 1st | N | 541 | 183 | 179 | 497 | 1,028 | 2,214 |
| | | P | 24.44% | 8.27% | 8.08% | 22.45% | 46.43% | 100.00% |
| | 2nd | N | 97 | 172 | 467 | 541 | 1,603 | 2,464 |
| | | P | 3.94% | 6.98% | 18.95% | 21.96% | 65.06% | 100.00% |
| | 3rd | N | 2,047 | 227 | 114 | 149 | 1,961 | 4,212 |
| | | P | 48.60% | 5.39% | 2.71% | 3.54% | 46.56% | 100.00% |
| | Total | N | 2,685 | 582 | 760 | 1,187 | 4,592 | 8,890 |
| | | P | 30.20% | 6.55% | 8.55% | 13.35% | 51.65% | 100.00% |
| **Unicredit Group** | 1st | N | 263 | 287 | 34 | 307 | 589 | 1,394 |
| | | P | 18.87% | 20.59% | 2.44% | 22.02% | 42.25% | 100.00% |

|  |  |  |  |  |  |  |  |
|---|---|---|---|---|---|---|---|
|  | 2nd | N | 79 | 195 | 20 | 109 | 391 | 724 |
|  |  | P | 10.91% | 26.93% | 2.76% | 15.06% | 54.01% | 100.00% |
|  | 3rd | N | 144 | 256 | 90 | 160 | 748 | 1,118 |
|  |  | P | 12.88% | 22.90% | 8.05% | 14.31% | 66.91% | 100.00% |
|  | Total | N | 486 | 738 | 144 | 576 | 1,728 | 3,236 |
|  |  | P | 15.02% | 22.81% | 4.45% | 17.80% | 53.40% | 100.00% |
| **Pirelli & C.** | 1st | N | 192 | 16 | 93 | 103 | 712 | 1,016 |
|  |  | P | 18.90% | 1.57% | 9.15% | 10.14% | 70.08% | 100.00% |
|  | 2nd | N | 179 | 23 | 2 | 11 | 995 | 1,192 |
|  |  | P | 15.02% | 1.93% | 0.17% | 0.92% | 83.47% | 100.00% |
|  | 3rd | N | 433 | 4 | 27 | 26 | 967 | 1,406 |
|  |  | P | 30.80% | 0.28% | 1.92% | 1.85% | 68.78% | 100.00% |
|  | Total | N | 804 | 43 | 122 | 140 | 2,674 | 3,614 |
|  |  | P | 22.25% | 1.19% | 3.38% | 3.87% | 73.99% | 100.00% |

Notes. N = number of tweets; P = percentage values.

As Table 2 shows, the highest volume of tweets was observed in the first collection week for all the leadership dimensions. Eni had the highest volume of tweets in each area, with more than 55% of all the tweets in the period: this was probably driven by the fact that at that time there was a journalistic inquiry about the company. Beside the unbalanced distribution of tweets among collection periods, we also notice that most of the tweets that refer to a leadership area pertain to the symbolic one, which accounts for more than 20% of the whole traffic volume. The fact that a relatively large proportion of tweets is classified as "none of the others" is because a significant part of the discourse is related to topics which are not directly connectable to leadership styles. What is more important in our study is to understand the relative proportions of tweets in the different leadership areas, to have a first understanding of the leadership profiles of the analyzed companies. Moreover, if a manager should discover that the discourse about his/her company on Twitter is completely unrelated to leadership styles, this would be an important warning.

Figure 3 shows the overall distribution of tweets in each leadership dimension for each company.

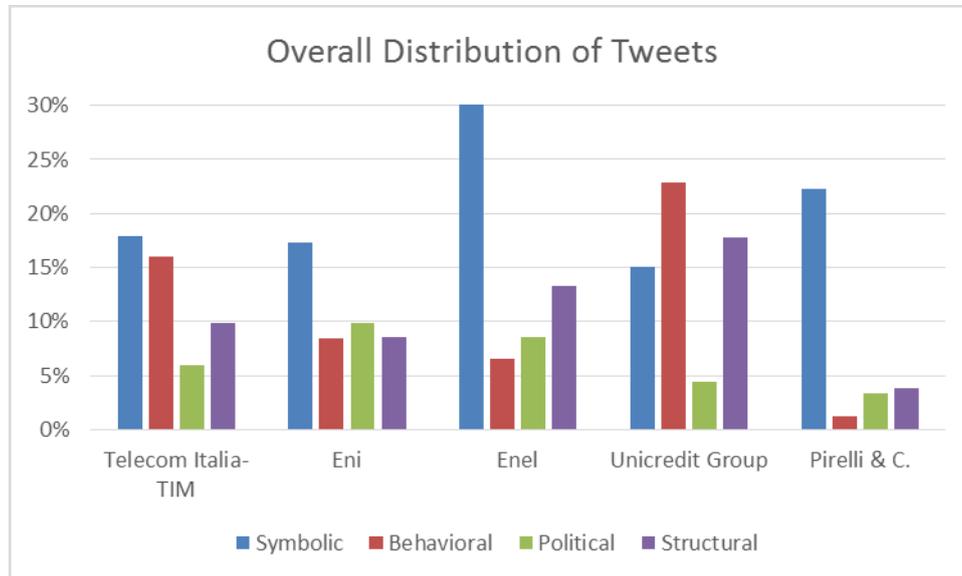

**Figure 3**. Overall distribution of tweets.

This study was neither conceived with the aim of presenting a ranking of the five companies, nor with the idea of proving a causal link between the results and specific facts or management policies. These companies offer different products and operate in different business sectors, so they are not directly comparable. Our main objective is to present a tool which can offer a quick assessment of perceived leadership styles to company managers.

Telecom Italia-TIM had its best results in the Symbolic and Behavioral dimensions, probably partially influenced by the fact that the company has a dedicated customer service on Twitter, which allowed direct interactions with customers, and by the rebranding campaign promoted to advertise the new company values. A lower performance was found in the Political and Structural areas; this might have been partially influenced by the frequent changes in the corporate governance that have been going on since 2014. These changes could have produced some fear in the employees due to the possibility of a downsizing of the company; moreover, such rotations in the governance could have generated a sense of uncertainty, even with respect to the possibility of actually realize the planned investments.

Results for Eni were partially influenced by the journalistic inquiry that involved the company and to which the company decided to answer publicly via Twitter. The company made a big effort to successfully control the discourse on Twitter. The best performance was in the Symbolic dimension followed by the Political one.

Enel had a relatively good performance in all the four dimensions of leadership, except for the Behavioral area. The area in which Enel had its best performance is the Symbolic dimension. This is in line with Fortune having acknowledged the company with the fifth position in the "Change the world" 2015 list. As regards the good performance in the Structural dimension, this could be partially justified with the fact that, at the beginning of 2016, the CEO of Enel Green Power (one of the subsidiaries of Enel) led the merger between the subsidiary and the parent company, to take advantage of the potential synergies of the two organizations.

Among the five organizations analyzed UniCredit Group is the one that had the lowest number of tweets. Results show that Unicredit had its best performance in the Behavioral area, followed by the Structural and Symbolic areas. This good positioning was obtained notwithstanding the layoff plans announced in September 2015, probably thanks to the social activities carried out by the Unicredit Foundation. The performance in the Political dimension was rather poor.

Lastly, Pirelli & C. showed relatively poor results in the Behavioral, Political and Structural areas. This seems to be at least partially influenced by the change in the ownership of the company in 2015, when it was acquired by the foreign company ChemChina. The only dimension in which the organization had a good performance is the Symbolic area; this is probably attributable to the fact that Pirelli is the exclusive supplier of major sport competitions, such as the Forumla One Championship. This has a strong communicative impact, consistent with the brand positioning, and enforces the trust of consumers in the high quality of the company products.

## Discussion and Conclusions

In this paper, we propose a text classification approach to evaluate perceived organizational leadership profiles, based on a SVM algorithm. We referred to the four leadership styles of the ten factor leadership model proposed by Barchiesi and La Bella (2007), which was combined with our methods to overcome some of the limitations of traditional surveys. We analyzed the dialogue with the stakeholders on Twitter and the leadership styles of five large multinational Italian companies.

Results show that the insights and the measures provided by our methodology can allow an almost real time assessment of perceived leadership styles and could help studying the impact of real events involving a company. The tool we developed can be very useful for firms which aim to understand their perceived leadership styles, as they emerge from the dialogue with the stakeholders and as they may appear to the generic social media user. Finally, an automated assessment of perceived leadership can also be helpful to compare the positioning of the company with respect to its competitors. This could be done analyzing a set of companies operating in the same industry. In this way, insights about a leadership repositioning strategy could be obtained, identifying the areas of possible improvement. Moreover, the analysis could be repeated over time, to draw a trajectory in the evolution of leadership profile.

This experimental study has several limitations, such as the monitoring of a single social media platform (Twitter) and the discussion of results that are referred to companies operating in different industries. We suggest replicating our experiment, possibly including other social media platforms – such as Facebook –, choosing a more homogeneous sample of firms and, if possible, distinguishing the tweets of the internal employees from those of the external stakeholders. We also suggest extending the observation period, to soften the effects of particular events which might happen just before or during the data collection. In this way, the assessment could be more pertinent to the actual feelings of the stakeholders, without being biased by exceptional events. For a more extensive assessment of perceived leadership styles, one could also take into account seasonal trends. However, the objective of our case study was not to provide a complete evaluation of company profiles, for instance for a year, but to illustrate a method which can provide an almost real time assessment, for specific events or timeframes. Lastly, our SVM classifier could be improved including the analysis of the sentiment of the language used in the tweets. Such a choice could offer a new element to help distinguishing between positive and negative tweets affecting each leadership area. It would also be possible to replicate the study looking at the internal communication of employees and leaders, for instance considering email messages instead of tweets, in order combine text mining with other leadership performance indicators (Tashiro et al., 2012).

# References

Aggarwal, C. C., & Zhai, C. (2012). A survey of text clustering algorithms. In C. C. Aggarwal, & C. Zhai, *Mining text data* (p. 77-128). New York, NY: Springer.Apte, C., Damerau, F., & Weiss, S. M. (1994). Automated learning of decision rules for text categorization. *ACM Transactions on Information Systems*, *12*(3), 233-251.

Baeza-Yates, R. A., & Ribeiro-Neto, B. (2011). *Modern Information Retrieval the Concepts and Technology Behind Search* (2nd ed.). Harlow, England: Pearson Education Ltd.

Banerjee, S., Chua, A. Y. K., & Kim, J. J. (2017). Don't be deceived: Using linguistic analysis to learn how to discern online review authenticity. *Journal of the Association for Information Science and Technology*, in press. http://doi.org/10.1002/asi.23784

Barchiesi, M. A., & La Bella, A. (2007). Leadership Styles of World's most Admired Companies: A Holistic Approach to Measuring Leadership Effectiveness. *Proceedings of 2007 International Conference on Management Science & Engineering* (*14th*) (p. 1437-1447). Harbin, P.R. China: Harbin Institute of Technology Press.

Barrett, D. J. (2006). Leadership Communication: A Communication Approach for Senior-Level Managers. In *Handbook of Business Strategy* (p. 385-390). Houston, Texas: Emerald Group Publishing.

Bass, B. M. (1985). *Leadership and performance beyond expectations*. New York, NY: The Free Press.

Bass, B. M. (1996). *A new paradigm of leadership: An inquiry into transformational leadership*. Alexandria, VA: U.S. Army Research Institute for the Behavioral and Social Sciences.

Bass, B. M. (2013). Forecasting Organizational Leadership: From Back (1967) to the Future (2034). In B. J. Avolio , F. J. Yammarino (Eds.) *Transformational and Charismatic Leadership: The Road Ahead 10th Anniversary Edition* (pp. 439-448). Bingley, UK: Emerald Group Publishing Limited.

Bass, B. M., & Avolio, B. J. (1990). *Multifactor leadership questionnaire*. Palo Alto, CA: Consulting Psychologists Press.

Bass, B. M., Avolio, B. J., & Goodheim, L. (1987). Biography and the Assessment of Transformational Leadership at the World Class Level. *Journal of management*, *13*(1), 7-19.

Bass, B. M., Avolio, B. J., Jung, D. I., & Berson, Y. (2003). Predicting Unit Performance by Assessing Transformational and Transactional Leadership. *Journal of Applied Psychology*, *88*(2), 207-218.

Behrendt, P., Matz, S., & Göritz, A. S. (2017). An integrative model of leadership behavior. *The Leadership Quarterly, 28*(1), 229–244.

Burns, J. M. (1978). *Leadership*. New York, NY: Harper & Row.

Conger, J. A. (1989). *The charismatic leader: Behind the mystique of exceptional leadership*. San Francisco, CA: Jossey-Bass.

Conger, J. A., & Kanungo, R. (1998). *Charismatic leadership in organizations*. Thousand Oaks, CA: Sage Publications.

Conger, J. A., & Kanungo, R. N. (1987). Toward a behavioral theory of charismatic leadership in organizational settings. *Academy of Management Review*, *12*, 637-647.

Conger, J. A., & Kanungo, R. N. (1994). Charismatic leadership in organizations: Perceived behavioral attributes and their measurement. *Journal of Organizational Behavior*, *15*(5), 439-452.

Conger, J. A., Kanungo, R., Menon, S. T., & Mathur, P. (1997). Measuring charisma: Dimensionality and validity of the Conger-Kanungo scale of charismatic leadership. *Canadian Journal of Administrative Sciences*, *14*, 290-302.

Day, D. V., Fleenor, J. W., Atwater, L. E., Sturm, R. E., & McKee, R. A. (2014). Advances in leader and leadership development: A review of 25 years of research and theory. *Leadership Quarterly, 25*(1), 63–82.

De Mauro, A., Greco, M., & Grimaldi, M. (2016). A formal definition of Big Data based on its essential features. *Library Review*, *65*(3), 122–135.

Elshendy, M., Fronzetti Colladon, A., Battistoni, E., & Gloor, P. A. (2017). Using Four Different Online Media Sources to Forecast the Crude Oil Price. *Journal of Information Science*, in press. http://doi.org/10.1177/0165551517698298

Farzindar, A. (2012). Industrial perspectives on social networks. *EACL 2012-Workshop on Semantic Analysis in Social Media, Vol. 2*.

Farzindar, A., & Wael, K. (2015). A survey of techniques for event detection in twitter. *Computational Intelligence*, *31*(1), 132-164.

Grubera, D. A., Smerekb, R. E., Thomas-Huntc, M. C., & Jamesd, E. H. (2015). The real-time power of Twitter: Crisis management and leadership in an age of social media. *Business Horizons, 58*(2), 163–172.

Hastie, T., Tibshirani, R., & Friedman, J. (2009). *The Elements of Statistical Learning: Data Mining, Inference, and Prediction* (2nd ed.). Stanford, CA.: Springer.

Hogenboom, F., Frasincar, F., Kaymak, U., & De Jong, F. (2011). An overview of event extraction from text. In M. Van Erp, W. R. Van Hage, L. Hollink, A. Jameso, & R. Troncy (Eds.), *Workshop on Detection, Representation, and Exploitation of Events in the Semantic Web* (*DeRiVE2011*) *at Tenth International Semantic Web Conference* (*ISWC 2011*). *779*, p. 48-57. Koblenz, Germany: CEUR-WS.org.

Hollerit, B., Kröll, M., & Strohmaier, M. (2013). Towards Linking Buyers and Sellers: Detecting Commercial Intent on Twitter. *Proceedings of the 22nd International Conference on World Wide Web – WWW'13 Companion* (p. 629–632). Rio de Janeiro, Brazil: ACM Press.

House, R. J. (1977). A theory of charismatic leadership. In J. G. Hunt, & L. L. Larson, *Leadership: The cutting edge*. Carbondale, IL: Southern Illinois University.

Huffaker, D. (2010). Dimensions of Leadership and Social Influence in Online Communities. *Human Communication Research, 36*(4), 593–617.

Jansen, B. J., Zhang, M., Sobel, K., & Chowdury, A. (2009). Twitter power: Tweets as electronic word of mouth. *Journal of the American Society for Information Science and Technology*, *60*(11), 2169–2188.

Jiang, L., Yu, M., Zhou, M., Liu, X., & Zhao, T. (2011). Target-dependent Twitter sentiment classification. *Proceedings of the 49th Annual Meeting of the Association for Computational Linguistics* (p. 151–160). Portland: Association for Computational Linguistics.

Jivani, A. G. (2011). A Comparative Study of Stemming Algorithms. *International Journal of Computer Technology and Applications*, *2*(6), 1930–1938.

Jurafsky, D., & Martin, J. H. (2009). *Speech and Language Processing: An Introduction to Natural Language Processing, Computational Linguistics, and Speech Recognition* (2nd ed.). UpperSaddle River, NJ: Prentice Hall.

Lee, C., & Lee, G. G. (2006). Information gain and divergence-based feature selection for machine learning-based text categorization. *Information processing & management*, *42*(1), 155-165.

Lewis, D. D. (1998). Naive (Bayes) at forty: The independence assumption in information retrieval. *Proceedings of the 10th European conference on machine learning* (p. 4-15). Springer Berlin Heidelberg.

Lovejoy, K., Waters, R. D., & Saxton, G. D. (2012). Engaging stakeholders through Twitter: How nonprofit organizations are getting more out of 140 characters or less. *Public Relations Review*, *38*(2), 313-318.

Lowe, K. B., Kroeck, K. G., & Sivasubramaniam, N. (1996). Effectiveness correlates of transformational and transactional leadership: A meta-analytic review of the MLQ literature. *The Leadership Quarterly*, *7*(3), 385-415.

Manning, C. D., & Schütze, H. (1999). *Foundations of statistical natural language processing*. Cambridge, MA: MIT press.

Murphy, K. P. (2012). *Machine learning: a probabilistic perspective*. Cambridge, MA: MIT press.

Ounis, I. (2009). Inverse Document Frequency. In L. Liu & M. T. Özsu (Eds.), *Encyclopedia of Database Systems* (pp. 1570–1571). New York, NY: Springer US.

Ozmutlu, H. C., Cavdur, F., & Ozmutlu, S. (2008). Cross-validation of neural network applications for automatic new topic identification. *Journal of the American Society for Information Science and Technology, 59*(3), 339–362

Pak, A., & Paroubek, P. (2010). Twitter as a corpus for sentiment analysis and opinion mining. *Proceedings of the Seventh International Conference on Language Resources and Evaluation* (*LREC'10*). Valletta, Malta: European Language Resources Association (ELRA).

Pedregosa, F., Varoquaux, G., Gramfort, A., Michel, V., Thirion, B., Grisel, O., … Duchesnay, É. (2011). Scikit-learn: Machine Learning in Python. *Journal of Machine Learning Research*, *12*, 2825–2830.

Perkins, J. (2014). *Python 3 Text Processing With NLTK 3 Cookbook*. Birmingham, UK: Packt Publishing.

Podsakoff, P. M., MacKenzie, S. B., Lee, J. Y., & Podsakoff, N. P. (2003). Common method biases in behavioral research: a critical review of the literature and recommended remedies. *The Journal of Applied Psychology*, *88*(5), 879–903.

Podsakoff, P. M., MacKenzie, S. B., Moorman, R. H., & Fetter, R. (1990). Transformational Leader Behaviors and Their Effects on Followers' Trust in Leader, Satisfaction, Organizational Citizenship Behaviors. *Leadership Quarterly*, *1*(2), 107-142.

Rybalko, S., & Seltzer, T. (2010). Dialogic communication in 140 characters or less: How Fortune 500 companies engage stakeholders using Twitter. *Public Relations Review*, *36*(4), 336-341.

Seltzer, T., & Mitrook, M. A. (2007). The dialogic potential of weblogs in relationship building. *Public Relations Review*, *33*(2), 227-229.

Shamir, B., House, R. J., & Arthur, M. B. (1993). The motivational effects of charismatic leadership: A self-concept theory. *Organization Science*, *4*(4), 577-594.

Shamir, B., Zakay, E., Breinin, E., & Popper, M. (1998). Correlates of charismatic leader behavior in military units: Subordinates' attitudes, unit characteristics, and superiors' appraisals of leader performace. *Academy of Management Journal*, *41*, 387-409.

Stogdill, R. M. (1974). *Handbook of leadership: A survey of the literature*. New York, NY: Free Press.

Stringhini, G., Kruegel, C., & Vigna, G. (2010). Detecting spammers on social networks. In *Proceedings of the 26th Annual Conference on Computer Security Applications - ACSAC '10* (pp. 1–9). New York, NY: ACM Press.

Tashiro, H., Lau, A., Mori, J., Fujii, N., & Kajikawa, Y. (2012). E-mail networks and leadership performance. *Journal of the American Society for Information Science and Technology, 63*(3), 600–606.

Tichy, N. M., & Devanna, M. A. (1986). *The transformational leader*. New York, NY: Wiley.

Trice, H. M., & Beyer, J. M. (1986). Charisma and its routinization in two social movement organizations. *Research in Organizational Behavior*, *8*, 113-164.

Van Riel, C. B., & Fombrun, C. J. (2007). *Essentials of corporate communication: Implementing practices for effective reputation management*. Routledge.

Vapnik, V. (2013). *The nature of statistical learning theory*. Springer Science & Business Media.

Waldman, D. A., Bass, B. M., & Einstein, W. O. (1987). Leadership and outcomes of performance appraisal processes. *Journal of Occupational Psychology*, *60*, 177-186.

Wang, Z. Q., Sun, X., Zhang, D. X., & Li, X. (2006). An optimal SVM-based text classification algorithm. *Proceedings of the Fifth International Conference on Machine Learning and Cybernetics* (p. 1378-1381). IEEE.

Yang, Y. (1999). An evaluation of statistical approaches to text categorization. *Information retrieval*, *1*(1-2), 69-90.Yukl, G. (1989). Managerial Leadership: A Review of Theory and Research. *Journal of Management*, *15*(2), 251-289.

Yukl, G. (1999). An evaluation of conceptual weaknesses in transformational and charismatic leadership theories. *Leadership Quarterly*, *10*(2), 285-305.

Zheng, X., Zeng, Z., Chen, Z., Yu, Y., & Rong, C. (2015). Detecting spammers on social networks. *Neurocomputing*, *159*(1), 27–34.